\documentclass[
nonacm,
authorversion,
twocolumn
]{ceurart}

\usepackage{listings}
\usepackage[ruled,vlined,linesnumbered]{algorithm2e}
\usepackage{amsmath} % For math symbols
\usepackage{booktabs} % For professional tables
\usepackage{siunitx} % For plus-minus formatting
\usepackage{caption}
\usepackage{microtype}
\usepackage{graphicx}
\usepackage{algorithmic}
\usepackage{pifont}
\usepackage{subfigure}
\usepackage{float}
\usepackage{rotating}
\usepackage{hyperref}  % Load last to avoid conflicts

\begin{document}

\copyrightyear{2024}
\copyrightclause{Copyright for this paper by its authors.
  Use permitted under Creative Commons License Attribution 4.0
  International (CC BY 4.0).}

\conference{ROEGEN@RecSys'24: The 1st Workshop on Risks, Opportunities, and Evaluation of Generative Models in Recommender Systems, Co-located with ACM RecSys in Bari, Italy, October 2024}

\title{Generative Diffusion Models for Sequential Recommendations}

%\author[1,2]{Dmitry S. Kulyabov}[%
%orcid=0000-0002-0877-7063,
%email=kulyabov-ds@rudn.ru,
%url=https://yamadharma.github.io/,
%]
%\cormark[1]
%\address[1]{Peoples' Friendship University of Russia (RUDN University),
%  6 Miklukho-Maklaya St, Moscow, 117198, Russian Federation}
%\address[2]{Joint Institute for Nuclear Research,
%  6 Joliot-Curie, Dubna, Moscow region, 141980, Russian Federation}
\author[1]{Sharare Zolghadr}[
email=sharare.zolghadr@studenti.unipd.it]
\address[1]{University of Padua, Padua, Italy}

\author[2]{Ole Winther}[
email=olwi@dtu.dk]

\author[2]{Paul Jeha}[
email=pauje@dtu.dk]
\address[2]{Technical University of Denmark, Copenhagen, Denmark}

%\author[4]{Manfred Jeusfeld}[%
%orcid=0000-0002-9421-8566,
%email=Manfred.Jeusfeld@acm.org,
%url=http://conceptbase.sourceforge.net/mjf/,
%\address[4]{University of Skövde, Högskolevägen 1, 541 28 Skövde, Sweden}

%\cortext[1]{Corresponding author.}
%\fntext[1]{These authors contributed equally.}

\begin{abstract}
Generative models such as Variational Autoencoders (VAEs) and Generative Adversarial Networks (GANs) have shown promise in sequential recommendation tasks. However, they face challenges, including posterior collapse and limited representation capacity. The work by  \citeauthor{li2023diffurec}, \citeyear{li2023diffurec} is a novel approach that leverages diffusion models to address these challenges by representing item embeddings as distributions rather than fixed vectors. This approach allows for a more adaptive reflection of users' diverse interests and various item aspects. During the diffusion phase, the model converts the target item embedding into a Gaussian distribution by adding noise, facilitating the representation of sequential item distributions and the injection of uncertainty. An Approximator then processes this noisy item representation to reconstruct the target item. In the reverse phase, the model utilizes users' past interactions to reverse the noise and finalize the item prediction through a rounding operation. This research introduces enhancements to the DiffuRec architecture, particularly by adding offset noise in the diffusion process to improve robustness and incorporating a cross-attention mechanism in the Approximator to better capture relevant user-item interactions. These contributions led to the development of a new model, DiffuRecSys, which improves performance. Extensive experiments conducted on three public benchmark datasets demonstrate that these modifications enhance item representation, effectively capture diverse user preferences, and outperform existing baselines in sequential recommendation research.
\end{abstract}


\maketitle

\section{Introduction}

Recommender systems are algorithms that suggest items to users by analyzing various forms of input data. Their primary goal is to enhance the customer experience through personalized recommendations, often based on prior implicit feedback. These systems track user behaviors, such as purchase history, viewing habits, and browsing activity, to model user preferences. Sequential Recommendation, a specific type of recommendation, is particularly relevant for applications where user behavior is naturally sequential. It focuses on predicting the next item a user will interact with by considering the order of previous interactions. 

Mainstream solutions to Sequential Recommendation (SR) \cite{kang2018self} represent items with fixed vectors, which have a limited ability to capture the latent aspects of items and the diversity of user preferences. Generative models like Generative Adversarial Networks (GANs) \cite{goodfellow2020generative} and Variational Auto-Encoders (VAEs) \cite{kingma2013auto} have been widely applied in personalized recommendations, using adversarial training and encoder-decoder architectures, respectively, to model user behavior and preferences. However, Diffusion Models have shown significant advantages over GANs and VAEs, such as greater stability and higher generation quality in various tasks.

Diffusion Models (DMs) \cite{ho2020denoising, sohl2015deep, cai2020learning} have achieved state-of-the-art results in image synthesis tasks \cite{cai2020learning, ho2022cascaded, ho2022video, ramesh2022hierarchical, rombach2022high}. These models alleviate the trade-off between stability and quality by gradually corrupting images in a forward process and iteratively learning to reconstruct them. DMs progressively corrupt $\mathbf{x}_0$ with random noise and then recover $\mathbf{x}_0$ from the corrupted state $\mathbf{x}_T$ step by step. This forward process creates a tractable posterior \cite{sohl2015deep}, enabling the iterative modeling of complex distributions through flexible neural networks in the reverse generation process.

The objectives of recommender models align well with DMs, as recommendation essentially involves inferring future interaction probabilities from noisy historical interactions, where noise represents false-positive and false-negative items \cite{sato2020unbiased, wang2021denoising}. This makes DMs a promising approach for accurately modeling complex interaction patterns with strong representational ability. Despite their success in other domains, applying diffusion models to recommender systems remains underexplored. 

We further explore diffusion models for sequential recommendation (SR) by extending the method introduced by \citeauthor{li2023diffurec} (\citeyear{li2023diffurec}). Our work proposes significant enhancements to the existing architecture, resulting in a new model, \textbf{DiffuRecSys}\footnote{https://youtu.be/bEpDfAAGL2I}. Specifically, our contributions are as follows:

\begin{itemize}
    \item \textbf{Enhancing the Diffusion Recommender Model:} We incorporate cross-attention mechanisms within the Approximator of the model architecture. The model isn't just learning temporal dependencies (the sequential order of items) but also more complex relationships between past interactions and the target item by focusing on the most relevant past events.
    
    \item \textbf{Incorporation of Offset Noise:} We introduce offset noise into the diffusion process to increase model robustness and effectively handle variability in user interactions.
    
    \item \textbf{Comprehensive Experimental Validation:} We conduct extensive experiments across three datasets under various settings, demonstrating improvements of DiffuRec with our extensions over standard baselines.
\end{itemize}

\section{Related Works}
\subsection{Recommender Systems}
In the era of information overload, recommender systems are essential for filtering relevant information and have been widely adopted in e-commerce, online news, and social media. A core aspect of these systems is modeling user preferences based on past interactions. Traditional recommender systems are generally categorized into three main types: content-based filtering \citep{lops2011content}, collaborative filtering \citep{ linden2003amazon, sarwar2001item}, and hybrid methods \citep{burke2002hybrid}. However, while conventional approaches focus on static user preferences, sequential recommender systems account for the temporal dynamics of user behavior. These systems recognize that user preferences evolve over time, and the order of interactions provides critical context for making accurate recommendations. Early sequential recommendation models utilized Markov chains to capture item transition probabilities. These models could make recommendations based on recent interactions by modeling the likelihood of a user moving from one item to another. However, Markov chains often struggle to capture long-term dependencies due to their limited memory capacity \cite{rendle2010factorizing}.

Recurrent Neural Networks (RNNs) have been widely adopted for sequential recommendation tasks due to their ability to model temporal sequences. Popular variants, such as Long Short-Term Memory (LSTM) and Gated Recurrent Units (GRUs) effectively capture both short-term and long-term dependencies in user behavior. These models have significantly improved recommendation accuracy by leveraging sequential patterns in user interactions \cite{hidasi2015session}.

Attention mechanisms have further advanced sequential recommendation systems by enabling models to focus on the most relevant parts of the interaction history. Self-attention models, such as the Transformer, effectively capture complex dependencies within sequences, leading to state-of-the-art performance in various sequential recommendation tasks \cite{kang2018self}.

\subsection{Generative Models}
In the realm of generative modeling, the primary objective is to approximate an unknown distribution \( p_{\text{data}}(x) \) using a model distribution \( p_{\text{model}}(x) \). This is typically achieved by defining \( p_{\text{model}}(x; \theta) \), where \( \theta \) represents parameters that are adjusted to minimize the divergence between \( p_{\text{data}} \) and \( p_{\text{model}} \). A prevalent method for achieving this is maximum likelihood estimation, which minimizes the Kullback-Leibler (KL) divergence between the true and model distributions.

Deep generative models have recently demonstrated high-quality sampling across various data modalities. Models such as Generative Adversarial Networks (GANs), autoregressive models, and Variational Autoencoders (VAEs) have shown impressive capabilities in generating complex data types, including images and audio. This progress has sparked significant interest in adapting these models for generative recommender systems. By modeling the underlying generative process, such systems infer user interaction probabilities for items that have not been interacted with. Typically, these models assume that latent factors, such as user preferences, drive users' interactions with items (e.g., clicks). 

Generative recommender models \citep{wang2017irgan} can be broadly categorized into two types. First, GAN-based models employ a generator to estimate user interaction probabilities and rely on adversarial training for optimization \cite{jin2020sampling}. However, adversarial training can be unstable, which may lead to inconsistent performance. Second, VAE-based models use an encoder to approximate the posterior distribution over latent factors and aim to maximize the likelihood of observed interactions \cite{ma2019learning}. Although VAEs generally outperform GANs in recommendation tasks, they face a trade-off between tractability and representation power \citep{liang2018variational}. Simple encoders may struggle to capture heterogeneous user preferences, while more complex models may result in intractable posterior distributions \cite{kingma2016improved}. Ongoing research and innovation are crucial for advancing generative modeling and enhancing recommendation systems.

\section{Problem Statement}
In sequential recommendation, let \( U \) represent the set of users and \( I \) the set of items. For each user \( u \in U \), we organize the items that the user has interacted with in chronological order, forming a sequence \( S_u = [i^u_1, i^u_2, \ldots, i^u_{N-1}, i^u_N] \), where \( i^u_n \in I \) denotes the item that user \( u \) interacted with at the \( n \)-th timestamp, and \( N \) represents the maximum sequence length. The task of sequential recommendation is to predict the next item that user \( u \) is likely to interact with. During training, the model learns to maximize the probability of recommending the target item \( i_N \) based on the user's previously interacted items \( [i^u_1, i^u_2, \ldots, i^u_{N-1}] \). This probability is expressed as \( p(i^u_N \mid i^u_1, i^u_2, \ldots, i^u_{N-1}) \). During inference, the model predicts the probability of recommending the next item \( i_{N+1} \) based on the entire interaction sequence \( [i^u_1, i^u_2, \ldots, i^u_{N}] \), which can be written as \( p(i^u_{N+1} \mid i^u_1, i^u_2, \ldots, i^u_{N-1}, i^u_N) \). The ultimate goal of sequential recommendation is to generate a ranked list of items as candidates for the next item that user \( u \) is most likely to prefer.

\section{Preliminaries}
In this section, we provide a brief introduction to diffusion models and sequential recommendations. Subsequently, in Section \ref{sec:5} (Model Architecture), we will explain the methodology for adapting diffusion models to integrate with recommender systems.

\subsection{Diffusion Model}
\label{sec:4.2.1}
Deep diffusion probabilistic models can be briefly described as hierarchical variational autoencoders (VAEs) with a bottom-up path defined by a diffusion process (e.g., Gaussian diffusion) and a top-down path parameterized by deep neural networks (DNNs), representing a reversed diffusion process. Diffusion models \cite{sohl2015deep} are latent variable models characterized by the distribution:
\begin{equation}
\begin{aligned}
&p_{\theta}(x_0) := \int p_{\theta}(x_{0:T}) \, dx_{1:T},
\end{aligned}
\end{equation}
where \( x_1, \ldots, x_T \) are latent variables with the same dimensionality as the data \( x_0 \sim q(x_0) \). The joint distribution \( p_{\theta}(x_{0:T}) \) represents the reverse process and is defined as a Markov chain with learned Gaussian transitions starting from \( p(x_T) = \mathcal{N}(x_T; 0, I) \):
\begin{equation}
\begin{aligned}
& p_{\theta}(x_{0:T}) := p(x_T) \prod_{t=1}^{T} p_{\theta}(x_{t-1} \mid x_t), \\
& p_{\theta}(x_{t-1} \mid x_t) := \mathcal{N}(x_{t-1}; \mu_{\theta}(x_t, t), \Sigma_{\theta}(x_t, t)).
\end{aligned}
\label{eq:1}
\end{equation}

What distinguishes diffusion models from other latent variable models is that the approximate posterior \( q(x_{1:T} \mid x_0) \), known as the forward process or diffusion process, is fixed as a Markov chain that gradually adds Gaussian noise to the data according to a variance schedule \( \beta_1, \ldots, \beta_T \):
\begin{equation}
\begin{aligned}
& q(x_{1:T} \mid x_0) := \prod_{t=1}^{T} q(x_t \mid x_{t-1}), \\
& q(x_t \mid x_{t-1}) := \mathcal{N}(x_t; \sqrt{1 - \beta_t} x_{t-1}, \beta_t I).
\end{aligned}
\label{eq:2}
\end{equation}

Here, \( x_t \) is sampled from this Gaussian distribution, where \( \beta_t \) controls the amount of noise added at the \( t \)-th diffusion step, and \( I \) is the identity matrix. The value of \( \beta_t \) is determined by a predefined noise schedule \( \beta \), which specifies the noise amount for each step. Common noise schedules include square-root \cite{nichol2021improved}, cosine \cite{dhariwal2021diffusion}, and linear \cite{ho2020denoising}.

In the reverse phase, where the original representation \( x_0 \) is unknown, a deep neural network \( f_\theta (\cdot) \) (e.g., Transformer \cite{vaswani2017attention} or U-Net \cite{ronneberger2015u}) is typically used to estimate \( x_0 \). Given the original representation \( x_0 \) and the schedule \( \beta \), training \( f_\theta (\cdot) \) involves optimizing the variational lower bound (VLB) on the negative log-likelihood \cite{sohl2015deep}:

\begin{multline}
\mathbb{E}[-\log p_{\theta}(x_0)] \leq \mathbb{E}_q \left[ -\log \frac{p_{\theta}(x_{0:T})}{q(x_{1:T} \mid x_0)} \right] =\\
 \mathbb{E}_q \Bigg[ -\log p(x_T) -
 \sum_{t \geq 1} \log \frac{p_{\theta}(x_{t-1} \mid x_t)}{q(x_t \mid x_{t-1})} \Bigg] =: \mathcal{L}.
\label{eq:3}
\end{multline}

The forward process variances, $\beta_t$, can either be adjusted via parameterization \cite{kingma2013auto} or kept fixed as hyperparameters. The reverse process remains effective due to the use of Gaussian conditionals in $p_\theta(x_{t-1} \mid x_t)$, particularly when $\beta_t$ values are small \cite{sohl2015deep}. A notable property of the forward process is that it facilitates straightforward sampling of $x_t$ at any time step $t$. Using the notation $\alpha_t := 1 - \beta_t$ and $\bar{\alpha}_t := \prod_{s=1}^t \alpha_s$, we express:
\begin{equation}
q(x_t \mid x_0) = \mathcal{N}(x_t; \sqrt{\bar{\alpha}_t} x_0, (1 - \bar{\alpha}_t) I),
\end{equation}

Efficient training can be achieved by optimizing the random terms of $\mathcal{L}$ using stochastic gradient descent. To further enhance training stability, we can reduce variance by rewriting $\mathcal{L}$ \eqref{eq:3} as:
\begin{equation}
\begin{aligned}
\mathbb{E}_q \Bigg[
    & \underbrace{D_{KL}\left( q(x_T \mid x_0) \parallel p(x_T) \right)}_{\mathcal{L}_T} + \\
    & \sum_{t>1} \underbrace{D_{KL}\left( q(x_{t-1} \mid x_t, x_0) \parallel p_\theta(x_{t-1} \mid x_t) \right)}_{\mathcal{L}_{t-1}} \\
    & \quad - \underbrace{\log p_\theta(x_0 \mid x_1)}_{\mathcal{L}_0}
\Bigg],
\end{aligned}
\label{eq:6}
\end{equation}

Equation \eqref{eq:6} leverages KL divergence to compare $p_{\theta}(x_{t-1} \mid x_t)$ against the forward process posteriors, which are tractable when conditioned on $x_0$:
\begin{equation}
q(x_{t-1} \mid x_t, x_0) = \mathcal{N}(x_{t-1}; \tilde{\mu}_t(x_t, x_0), \tilde{\beta}_t I),
\end{equation}
where
\begin{equation}
\begin{split}
\tilde{\mu}_t(x_t, x_0) &:= \frac{\sqrt{\bar{\alpha}_{t-1}} \beta_t}{1 - \bar{\alpha}_t} x_0 + \frac{\sqrt{\alpha_t }(1 - \bar{\alpha}_{t-1})}{1 - \bar{\alpha}_t} x_t,
\\   \quad \tilde{\beta}_t &:=\frac{1 - \bar{\alpha}_{t-1}}{1 - \bar{\alpha}_t} \beta_t,
\end{split}
\end{equation}

Thus, all KL divergences in Equation \eqref{eq:6} are Gaussian comparisons, which can be calculated exactly using closed-form expressions, rather than high-variance Monte Carlo estimates.

The loss function in Equation \eqref{eq:6} is organized into three components: $L_t$, $L_{t-1}$, and $L_0$. Component $L_t$ aims to bring $q(x_t \mid x_0)$ close to a standard Gaussian distribution. Component $L_{t-1}$ minimizes the KL divergence between the forward process posterior and $p_{\theta}(x_{t-1} \mid x_t)$, which is generated by the deep neural network for the reverse process. The last Component $L_0$ applies negative log-likelihood for the final prediction. This variational lower bound (VLB) loss could lead to unstable model training. To address this, each $L_t$ can be further simplified to:
\begin{equation}
L_{t, \text{simple}} = \mathbb{E}_{t, x_0, \epsilon} \left[ \left\| \epsilon - \epsilon_{\theta} \left( \sqrt{\bar{\alpha_t}} x_0 + \sqrt{1 - \bar{\alpha_t}} \epsilon, t \right) \right\|^2 \right],
\label{eq:8}
\end{equation}
where $\epsilon \sim \mathcal{N} (0, I)$ is sampled from a standard Gaussian distribution for noise injection and diffusion, and $\epsilon_{\theta} (\cdot)$ serves as an approximator instantiated by a deep neural network, such as a Transformer or U-Net.

\subsection{Sequential Recommendation}
Let \( R \in \mathbb{R}^{m \times n} \) represent the user-item interaction matrix, where \( m \) is the number of users and \( n \) is the number of items. Each entry \( r_{ui} \) indicates the interaction (e.g., rating or click) between user \( u \) and item \( i \). Each user \( u \) has a sequence of interacted items \( S_u = [i_1, i_2, \ldots, i_{T=N_u}] \), where \( N_u \) is the length of the sequence. Items are embedded into a latent space using an embedding matrix \( E \in \mathbb{R}^{n \times d} \), where \( d \) is the dimensionality of the embedding. The embedding of item \( i \) is denoted as \( e_i = E[i] \).

RNNs are used to model the sequential nature of user interactions. For a sequence \( S_u \), the hidden state \( h_t \) at time step \( t \) is updated as
\begin{equation}
h_t = \text{RNN}(h_{t-1}, e_{i_t}),
\end{equation}
where \( e_{i_t} \) is the embedding of the item at time \( t \), and \( h_{t-1} \) is the hidden state from the previous time step.

The self-attention mechanism can be used to weigh the importance of each item in the sequence. The attention score \( \alpha_{tj} \) between item \( i_t \) and item \( i_j \) is computed as
\begin{equation}
\alpha_{tj} = \frac{\exp(e_{i_t}^\top e_{i_j})}{\sum_{k=1}^{T_u} \exp(e_{i_t}^\top e_{i_k})},
\end{equation}
where \( \exp \) denotes the exponential function.

The attended representation \( z_t \) is then
\begin{equation}
z_t = \sum_{j=1}^{T_u} \alpha_{tj} e_{i_j}.
\end{equation}

The model predicts the next item \( \hat{i}_{T_u+1} \) based on the final hidden state \( h_{T_u} \) or the attended representation \( z_{T_u} \) as
\begin{equation}
\hat{y} = W h_{T_u} + b,
\end{equation}
where \( W \) and \( b \) are learned parameters, and \( \hat{y} \) is the predicted score vector for the next item.

\section{Model Architecture}
\label{sec:5}
\begin{figure*}[!h]
    \centering
    \includegraphics[width=0.8\linewidth]{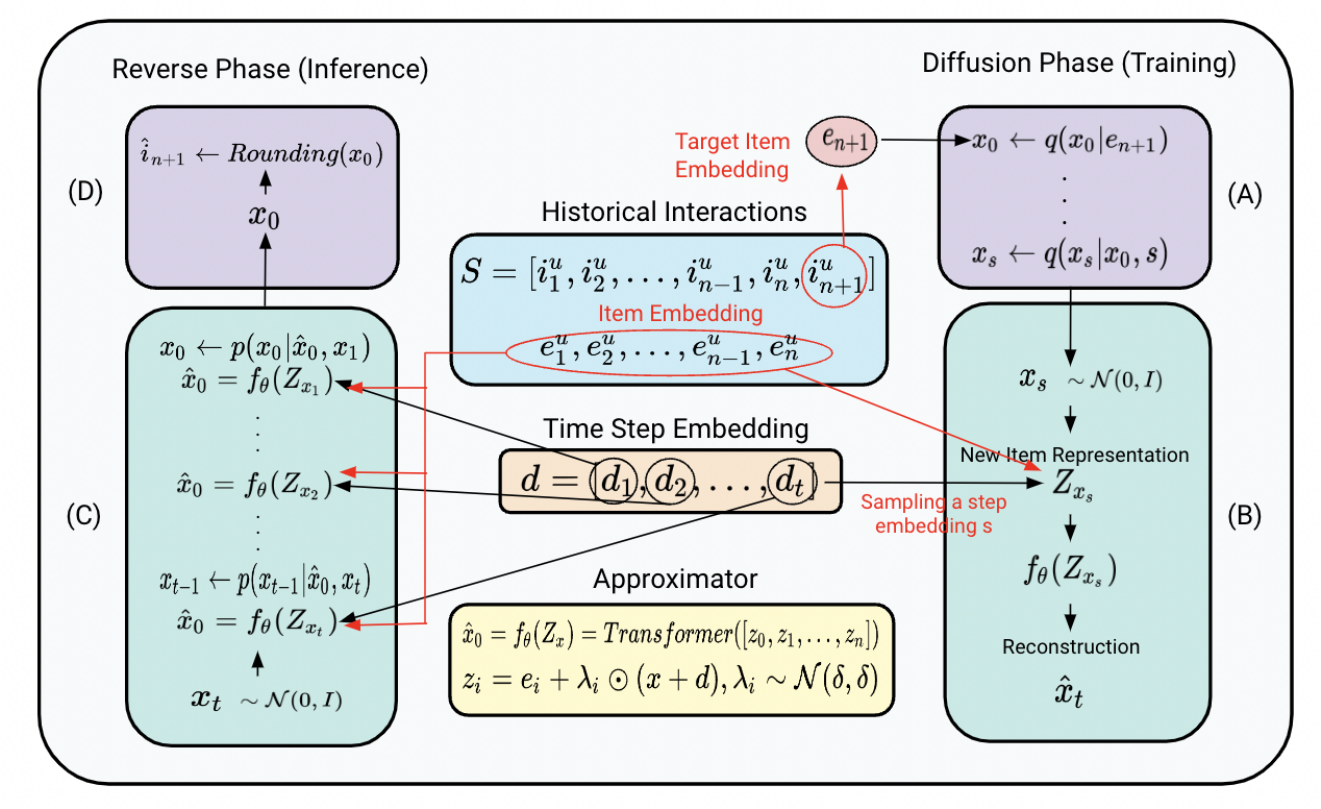}
    \caption{\small Overview of the diffusion process for sequential recommendation: (A) Injecting noise into the target item after $s$ diffusion steps; (B) Generating new item representation based on user history and the last target item; (C) Reverse phase for target item reconstruction; (D) Rounding phase to map the continuous target representation to discrete item indices.}
    \label{fig:Graph}
\end{figure*}

The DiffuRec framework is shown in Figure \ref{fig:Graph}. The input to the model is the embedding \( e_{n+1} \) of the target item \( i_{n+1} \). According to the diffusion model detailed in Section \ref{sec:4.2.1}, we aim to reverse this process to retrieve the target item from the historical sequence \( S \). To achieve this, we introduce noise into the target item embedding \( e_{n+1} \) via the diffusion process, which involves multiple samplings from Gaussian distributions (Equation \ref{eq:2}). The noised target item representation \( x_s \), which has undergone \( s \) diffusion steps, is considered as the distribution representation sampled from \( q(\cdot) \). Consequently, \( x_s \) can still encapsulate multiple latent aspects of the item.
Next, we use \( x_s \) to adjust the new representation of each historical item embedding in \( S \), leveraging the target item's guidance as additional semantic signals. The resulting representations \( Z_{x_s} = [z_1, z_2, \ldots, z_n] \) are then input into the approximator \( f_\theta(\cdot) \). The model is trained to ensure the reconstructed \( \hat{x}_0 \) from the approximator closely matches the target item embedding \( e_{n+1} \).
In the reverse phase, we start by sampling the noised target item representation \( x_t \) from a standard Gaussian distribution, \( x_t \sim \mathcal{N}(0, I) \). Similar to the diffusion process, the adjusted representations \( Z_{x_t} \) (obtained using \( x_t \)) are fed into the well-trained approximator \( f_\theta(\cdot) \) for \( \hat{x}_0 \) estimation. Following Equation \eqref{eq:1}, the estimated \( \hat{x}_0 \) and \( x_t \) are utilized to reverse \( x_{t-1} \) through \( p(\cdot) \). This iterative process continues until \( x_0 \) is reached. It is important to note that the reverse phase is stochastic, which models the uncertainty in user behavior. Finally, a rounding function maps the reversed continuous representation \( x_0 \) into discrete candidate item indices for predicting the target item \( \hat{i}_{n+1} \).
We present three versions of the Diffusion Recommender Model: 1) a baseline based on previous work \cite{li2023diffurec}, 2) a model with cross-attention in the approximator, and 3) a model with both offset noise and cross-attention. We compare the performance of all models to assess the effectiveness of each component.
\subsection{Approximator}
\label{sec:4.3.1}
The transformer is employed as the backbone of the approximator \( f_{\theta}(\cdot) \) to generate \( \hat{x}_0 \), leveraging its effectiveness in modeling sequential dependencies during both the diffusion and reverse phases:
\begin{equation}
\hat{x}_0 = f_{\theta}(Z_{x}) = \text{Transformer}([z_0, z_1, \ldots, z_n]).
\end{equation}

To enhance the representation of the user's historical interactions with the last target item, we employ cross-attention. Instead of summing the last target item and timestep embedding with the historical interactions embedding, as shown in Equation (\ref{eq:30}), we utilize cross-attention, as depicted in Equation (\ref{eq:31}):
\begin{equation}
z_i = e_i + \lambda_i \odot (x + d), \label{eq:30}
\end{equation}
\begin{equation}
z_i = \text{CrossAttention}(e_i, \lambda_i \odot (x + d)), \label{eq:31}
\end{equation}

In Equation (\ref{eq:31}), \( \odot \) denotes element-wise multiplication, and \( d \) represents the step embeddings that incorporate specific information for each diffusion and reverse step. In the diffusion phase, \( x \) is the noised target item embedding \( x_s \). In the reverse process, \( x \) refers to the reversed target item representation, i.e., \( x = x_s \) for \( s = t, t - 1, \ldots, 2, 1 \).

\begin{figure}[!h]
    \centering
    \includegraphics[width=1\columnwidth]{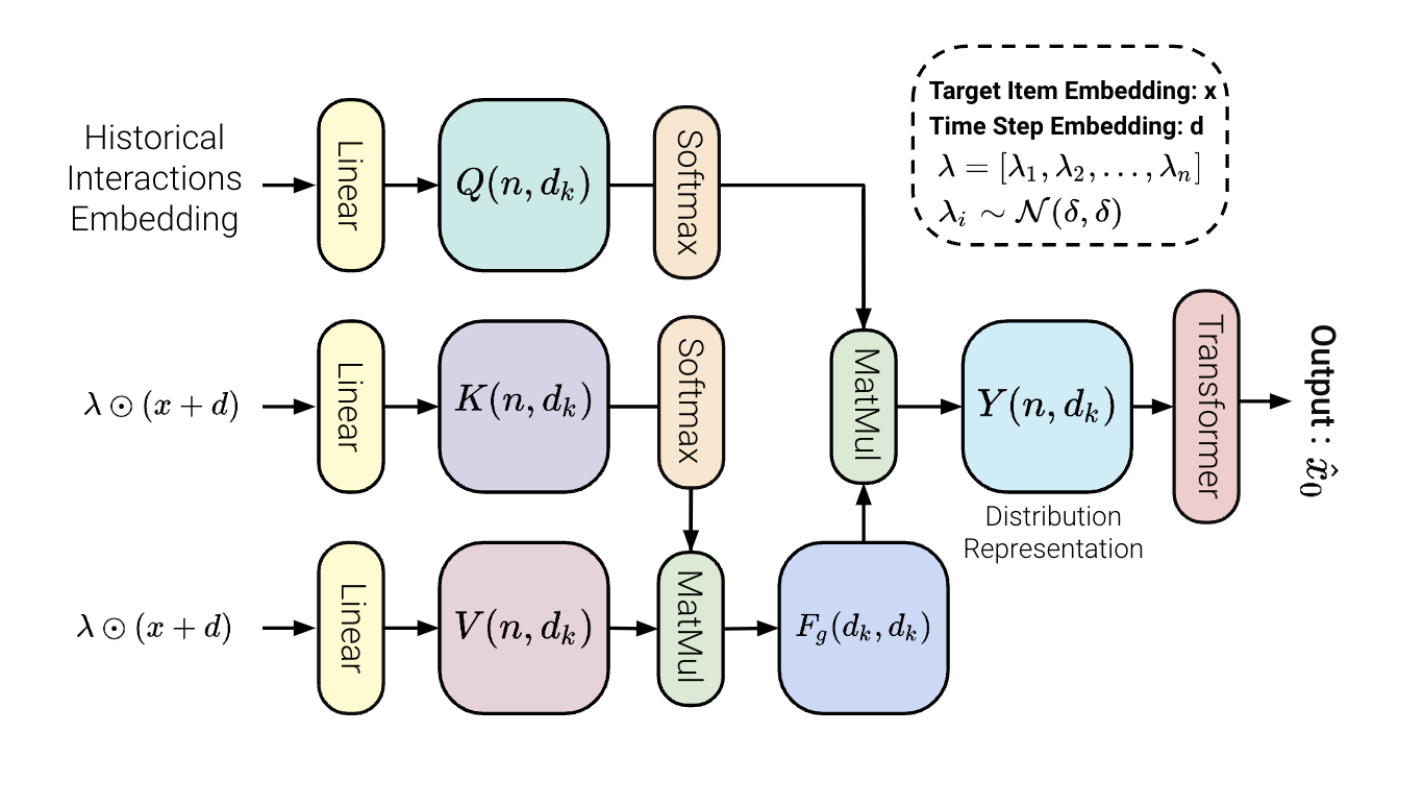}
    \caption{\small The architecture of approximation using cross-attention between the last target item and the user's historical interactions.}
    \label{fig:GraphA}
\end{figure}

Cross-attention allows the model to dynamically weigh the importance of historical interactions to the last target item, enabling a more nuanced aggregation of information. This improves the model's ability to capture relevant context and relationships, leading to better item representation, a multi-head attention Transformer is used to predict the next item in the sequence. According to Equation (\ref{eq:31}), \( \lambda_i \) is sampled from a Gaussian distribution, \( \lambda_i \sim \mathcal{N}(\delta, \delta) \), where \( \delta \) is a hyperparameter that defines both the mean and variance. During the diffusion phase, \( \lambda_i \) determines the level of noise injection.

In the diffusion phase, a step-index \( s \) is randomly sampled from a uniform distribution over the range \([1, t]\), where \( t \) represents the total number of steps. Conversely, in the reverse phase, \( s \) ranges from \( t \) to 1. Consequently, both the corresponding step embedding and the sequence item distribution are fed into the approximator for model training (see Equation (\ref{eq:30})). Ultimately, the representation \( h_n \) of item \( i_{n} \) generated by the final layer is used as \( \hat{x}_0 \).

\subsection{Diffusion Phase}
\label{sec:4.3.2}
The offset noise technique was implemented to enhance the robustness and diversity of generated items in a diffusion model. The conventional approach in diffusion models involves sampling noise from a standard normal distribution, \( \mathcal{N}(0, I) \), where the mean is zero and the variance is one. However, this approach can make the model sensitive to mean shifts in the input data, potentially limiting the diversity and quality of generated outputs.

To address this issue, an offset term was introduced in the noise distribution. Specifically, the noise was sampled from a distribution with a constant offset term, denoted as:
\begin{equation}
\epsilon_{\text{offset}} \sim \mathcal{N}(0.1 \delta_c, I),
\end{equation}
instead of the standard normal distribution. This offset is applied only during the training or forward process. During inference or the reverse process, noise is still sampled from a standard normal distribution. Here, \( \delta_c \) represents a constant value, and the scaling factor \( 0.1 \) adjusts the mean of the noise distribution.

During training, we randomly sample a diffusion step \( s \) for each target item, specifically \( s = \lfloor s' \rfloor \) with \( s' \sim U(0, t) \).
 \( x_s \) is generated via:
\begin{equation}
q(x_s | x_0, s) = \mathcal{N}(x_s; \sqrt{\bar{\alpha_t}}x_0, (1 - \bar{\alpha_s})I). \
 \end{equation}
We derive \( x_0 \) through one-step diffusion from the target item embedding:
\begin{equation}
q(x_0 | e_{n+1}) = \mathcal{N}(x_0; \sqrt{\bar{\alpha}_0} e_{n+1}, (1 - \alpha_0) I).
\end{equation} 
Utilizing the reparameterization trick with \( \epsilon \sim \mathcal{N}(0.1 \delta_c, I) \), \( x_s \) can be generated as follows:
\begin{equation}
x_s = \sqrt{\bar{\alpha}_s} x_0 + \sqrt{1 - \bar{\alpha}_s} \epsilon.
\end{equation}
The diffusion process or model training is detailed in Algorithm \ref{alg:1}.
\begin{algorithm}[h]
\small
\caption{Training (Diffusion Phase)}
\label{alg:1}
\KwIn{User interactions $\overline{X}$, Initial parameters $\theta$}
\KwOut{Optimized parameters $\theta$}
    Sample a batch $X \subseteq \overline{X}$\;
    \For{all $x_0 \in X$}{
        Sample $t \sim \mathcal{U}(1, T)$, $\epsilon \sim \mathcal{N}(0.1 \delta_c, I)$\;
        $x_t \leftarrow q(x_t | x_0)$, $\lambda_i \sim \mathcal{N}(\delta, \delta)$\;
        $(z_1, \ldots, z_n) \leftarrow (e_1 + \lambda_1 (x_s + d_s), \ldots, e_n + \lambda_n (x_s + d_s))$\;
        $\hat{x}_0 \leftarrow f_\theta(z_1, \ldots, z_n)$\;
        Update $\theta$ via $L_{\text{CE}}(\hat{x}_0, i_{n+1})$\;
    }
    \textbf{While} not converged \textbf{do} Repeat above steps\;
\end{algorithm}
\subsection{Reverse Phase}
In the reverse phase, our goal is to iteratively reconstruct the target item representation \( x_0 \) from a purely Gaussian noise \( x_t \). However, obtaining \( x_0 \) directly at each reverse step is infeasible. Thus, we utilize a well-trained approximator to generate \( \hat{x}_0 \) as an estimate of \( x_0 \), specifically \( x_0 = \hat{x}_0 \). The reverse phase is detailed in Algorithm \ref{alg:2}. Subsequently, following Equation \eqref{eq:1}, the reverse process proceeds by applying the reparameterization trick:
\begin{equation}
\hat{x}_0 = f_{\theta}(Z_{x_t}), \quad x_{t-1} \leftarrow p(x_{t-1} \mid \hat{x}_0, x_t),
\end{equation}

\begin{equation}
x_{t-1} = \tilde{\mu}_t (x_t, \hat{x}_0) + \tilde{\beta}_t \epsilon', 
\end{equation}
where 
\begin{equation}
\begin{aligned}
\tilde{\mu}_t (x_t, \hat{x}_0) &= \frac{\sqrt{\bar{\alpha}_{t-1}} \beta_t}{1 - \bar{\alpha}_t} \hat{x}_0 + \frac{\sqrt{\bar{\alpha}_t} (1 - \bar{\alpha}_{t-1})}{1 - \bar{\alpha}_t} x_t,
\\ \quad \tilde{\beta}_t &= \frac{1 - \bar{\alpha}_{t-1}}{1 - \bar{\alpha}_t}\beta_t.
\end{aligned}
\end{equation}

After generating \( x_{t-1} \) using \( \epsilon' \sim \mathcal{N}(0, I) \), repeat the reverse process until reaching \( x_0 \).
\begin{algorithm}[h]
\small
\caption{Inference (Reverse Phase)}
\label{alg:2}
\KwIn{Sequence $(i_1, \ldots, i_n)$, Target $x_t \sim \mathcal{N}(0, I)$}
\KwOut{Predicted item $i_{n+1}$}
    \For{$s = T, \ldots, 1$}{
        $\lambda_i \sim \mathcal{N}(\delta, \delta)$\;
        $(z_1, \ldots, z_n) \leftarrow (e_1 + \lambda_1 (x_s + d_s), \ldots, e_n + \lambda_n (x_s + d_s))$\;
        $\hat{x}_0 \leftarrow f_\theta(z_1, \ldots, z_n)$, $\epsilon' \sim \mathcal{N}(0, I)$\;
        $x_{s-1} \leftarrow \tilde{\mu}_s (\hat{x}_0, x_s) + \tilde{\beta}_s \cdot \epsilon'$\;
    }
    $i_{n+1} \leftarrow \text{Rounding}(x_0)$
\end{algorithm}
\subsection{Loss Function and Rounding}
Following previous works \cite{strudel2022self,mahabadi2023tess,han2022ssd}, we adopt the cross-entropy loss during the diffusion phase for model optimization as follows:

\begin{equation}
\hat{y}_i = \frac{\exp(\hat{x}_0 \cdot e_{n+1})}{\sum_{i \in I} \exp(\hat{x}_0 \cdot e_i)},
\end{equation}
\begin{equation}
L_{CE} = -\frac{1}{|U|} \sum_{u \in U} \log \hat{y}_{u},
\end{equation}

where \( \hat{x}_0 \) is reconstructed by the Transformer-based approximator, and \( \cdot \) denotes the inner product operation.

During the inference phase, the task is to map the reversed target item representation \( x_0 \) into the discrete item index space for the final recommendation. To achieve this, we compute the inner product between \( x_0 \) and all candidate item embeddings \( e_i \) and subsequently select the item index corresponding to the maximum value as the recommended result. This process can be formalized as follows:

\begin{equation}
i_{n+1} = \text{arg max}_{i \in I} \text{Rounding}(x_0) = x_0 \cdot e_i^T.
\end{equation}

\section{Experiment}
In this section, we conduct experiments on three real-world datasets to address the following research question: 

\textbf{RQ1}) Does modifying the baseline Diffusion Recommender Model (DiffuRec) enhance performance? 

\textbf{RQ2}) How does the proposed DiffRecSys framework compare to competitive methods? 

\textbf{RQ3}) What are the weaknesses of the DiffRecSys model, and what solutions can be proposed to address them?

\subsection{Dataset}
For the experiments conducted in this study, we utilized the RecBole framework \cite{zhao2021recbole}, a comprehensive and unified tool for benchmarking recommender systems. RecBole provides an extensive collection of preprocessed datasets, which facilitated consistent and efficient data preparation. Specifically, we employed RecBole to download and preprocess the required datasets,  available at \url{https://github.com/RUCAIBox/RecSysDatasets}, ensuring compatibility and comparability across various recommendation models. The standardized environment of RecBole, along with its support for various algorithms and evaluation metrics, enabled a fair and reliable assessment of model performance.
We conducted experiments on three public benchmark datasets. Two subcategories of the Amazon dataset (Beauty and Toys) were selected, both comprising user reviews of products. Additionally, we utilized the MovieLens 1M dataset (ML1M), a widely recognized benchmark dataset containing approximately 1 million user ratings on movies.

\subsection{Experiment Setup}
For the training, validation, and test splits, we adopt a sequential split approach. Specifically, for all datasets, given a sequence \( S = \{i_1, i_2, \ldots, i_n\} \), the most recent interaction (\( i_n \)) is used for testing, the penultimate interaction (\( i_{n-1} \)) is used for validation, and the earlier interactions (\( \{i_1, i_2, \ldots, i_{n-2}\} \)) are used for training.

To assess the performance of sequential recommender systems, we evaluate all models using HR@K (Hit Rate) and NDCG@K (Normalized Discounted Cumulative Gain). HR@K measures the proportion of times the true item is among the top-K recommendations, while NDCG@K evaluates the ranking quality of the top-K items, taking into account the position of the correct items. The experimental results are reported for the top-K list with \( K \) set to 5, 10, and 20. We treat all reviews or ratings as implicit feedback (i.e., user-item interactions) and chronologically organize them by their timestamps. The maximum sequence length is set to 100 to balance capturing sufficient historical information with computational efficiency for the MovieLens-1M, Amazon Beauty, and Amazon Toys datasets. These datasets exhibit considerable diversity in sequence lengths and dataset sizes, covering a broad range of real-world scenarios. Additionally, we filter out unpopular items and inactive users who have fewer than five associated interactions.

For training, we employ the Adam optimizer with an initial learning rate of 0.001, a weight decay of 0.0001, and a learning rate decay step that reduces the learning rate starting at epoch 30, with the total number of epochs set to 41. Transformer parameters are initialized using Xavier initialization, and the number of blocks is set to 4.

We use the standard Transformer architecture and configuration as described by \cite{vaswani2017attention}, which includes multi-head self-attention, a feed-forward network with ReLU activation, layer normalization, dropout, and residual connections. To enhance the model's representation capacity, we stack multiple Transformer blocks.

Both the embedding dimension and hidden state size are fixed at 128, with a batch size of 1024. Dropout rates for the Transformer block and item embeddings are set to 0.1 and 0.3, respectively, across all datasets. For the hyperparameter \( \lambda \), values are sampled from a Gaussian distribution with a mean of 0.001 and a standard deviation of 0.001. The total number of reverse steps \( t \) is set to 32. For the noise schedule \( \beta \), a truncated linear schedule is used where the noise variance decreases linearly over the steps. The baseline model and two other variants are evaluated across multiple experimental runs, typically more than five, with average results reported. Statistical significance is assessed using a Student's t-test.

\subsection{Main Results (\textbf{RQ1})}

To evaluate the efficacy of our proposed modifications, we conducted a series of experiments on three distinct datasets. The experimental setup was designed to compare the performance of the baseline model (DiffuRec \cite{li2023diffurec}) with two enhanced models: one incorporating cross-attention alone and another combining cross-attention with Offset Noise. Each experiment was repeated five times with different random seeds to ensure the robustness and reliability of the results. Our objective was to determine whether these modifications led to performance improvements.

Comprehensive details of the experimental configurations and the average outcomes for each model are provided in Tables~\ref{tab:1}, \ref{tab:2}, and \ref{tab:4}. 

\begin{table}[h]
  \centering
  \caption{\small Comparison of DiffuRec and its Variants}
  \label{tab:1}
  \small % Reduce the font size for the table
 % Reduce the font size for the caption
  \begin{tabular}{ccc}
    \toprule
    \textbf{Model Version} & \textbf{Cross Attention} & \textbf{Offset Noise} \\
    \midrule
    DiffuRec (Baseline) & - & - \\
    Modified Version 1 & \ding{51} & -  \\
    Modified Version 2 & \ding{51} & \ding{51} \\
    \bottomrule
  \end{tabular}
    \vspace{0.6cm}
    \centering
    \caption{\small Convergence Training Time in Seconds}
    \label{tab:2}
 % Small font for the caption
    \small % Reduce the font size for the table
    
    \begin{tabular}{lccc}
        \toprule
        \textbf{Model} & \textbf{Movielens} & \textbf{Toys} & \textbf{Beauty} \\
        \midrule
        Baseline & 65312.18 & 7458.10 & 8734.22 \\
        Modified V1 & 30612.66 & \textbf{4470.52} & 6776.10 \\
        Modified V2 & \textbf{27283.95} & 4623.85 & \textbf{6468.38} \\
        \bottomrule
    \end{tabular}
    
\end{table}

\begin{table}[h]
  \centering
  \caption{\small Average Performance and Standard Deviation Over Multiple Experiments: Modified Versions vs. DiffuRec (Baseline)}
  \label{tab:4}
  \setlength{\tabcolsep}{3pt} % Adjust column padding
  \small
  \resizebox{\columnwidth}{!}{
  \begin{tabular}{lcccccc}
    \toprule
    \textbf{Dataset} & \textbf{Metric} & \textbf{Modified V1} & \textbf{Modified V2} & \textbf{DiffuRec} \\
    \midrule
    \multirow{6}{*}{Beauty} 
    & HR@5 & \textbf{0.0669}$_{\pm 0.0018}$ & 0.0667$_{\pm 0.0014}$ & 0.0557 \\ 
    & HR@10 & 0.0974$_{\pm 0.0021}$ & \textbf{0.0980}$_{\pm 0.0012}$ & 0.0790 \\ 
    & HR@20 & 0.1399$_{\pm 0.0023}$ & \textbf{0.1400}$_{\pm 0.0019}$ & 0.1110 \\ 
    & NDCG@5 & 0.0458$_{\pm 0.0012}$ & \textbf{0.0458}$_{\pm 0.0010}$ & 0.0400 \\ 
    & NDCG@10 & 0.0556$_{\pm 0.0012}$ & \textbf{0.0559}$_{\pm 0.0009}$ & 0.0475 \\ 
    & NDCG@20 & 0.0663$_{\pm 0.0011}$ & \textbf{0.0665}$_{\pm 0.0010}$ & 0.0556 \\ 
    \midrule
    \multirow{6}{*}{Movielens-1M} 
    & HR@5 & 0.1940$_{\pm 0.0029}$ & \textbf{0.1957}$_{\pm 0.0035}$ & 0.1797 \\ 
    & HR@10 & 0.2829$_{\pm 0.0027}$ & \textbf{0.2843}$_{\pm 0.0043}$ & 0.2626 \\ 
    & HR@20 & 0.3947$_{\pm 0.0044}$ & \textbf{0.3958}$_{\pm 0.0039}$ & 0.3679 \\ 
    & NDCG@5 & 0.1319$_{\pm 0.0022}$ & \textbf{0.1319}$_{\pm 0.0017}$ & 0.1212 \\ 
    & NDCG@10 & \textbf{0.1607}$_{\pm 0.0012}$ & 0.1605$_{\pm 0.0019}$ & 0.1479 \\ 
    & NDCG@20 & \textbf{0.1888}$_{\pm 0.0080}$ & 0.1886$_{\pm 0.0013}$ & 0.1744 \\ 
    \midrule
    \multirow{6}{*}{Toys} 
    & HR@5 & \textbf{0.0692}$_{\pm 0.0025}$ & 0.0684$_{\pm 0.0018}$ & 0.0557 \\ 
    & HR@10 & \textbf{0.0986}$_{\pm 0.0035}$ & 0.0971$_{\pm 0.0019}$ & 0.0746 \\ 
    & HR@20 & \textbf{0.1371}$_{\pm 0.0043}$ & 0.1350$_{\pm 0.0010}$ & 0.0984 \\ 
    & NDCG@5 & \textbf{0.0458}$_{\pm 0.0010}$ & 0.0455$_{\pm 0.0013}$ & 0.0417 \\ 
    & NDCG@10 & \textbf{0.0553}$_{\pm 0.0012}$ & 0.0548$_{\pm 0.0013}$ & 0.0477 \\ 
    & NDCG@20 & \textbf{0.0650}$_{\pm 0.0010}$ & 0.0643$_{\pm 0.0013}$ & 0.0537 \\ 
    \bottomrule
  \end{tabular}
  }
\end{table}

\begin{table}[h]
  \centering
  \caption{\small Paired t-test Results between Variants and Baseline}
  \label{tab:3}
  \resizebox{\columnwidth}{!}{ % Resize the table to fit within the column width
  \begin{tabular}{cccccc}
    \toprule
    \textbf{Dataset} & \textbf{Compared Model} & \textbf{t-statistic} & \textbf{p-value} & \textbf{Avg. Diff.} \\
    \midrule
    \multirow{2}{*}{Beauty} 
    & Modified V2 & 3.9853 & 0.0105 & 1.3845\% \\ 
    & Modified V1 & 4.0182 & 0.0101 & 1.4029\% \\ 
    \midrule
    \multirow{2}{*}{Movielens-1M} 
    & Modified V2 & 6.8432 & 0.0010 & 1.6575\% \\ 
    & Modified V1 & 6.5228 & 0.0013 & 1.7198\% \\ 
    \midrule
    \multirow{2}{*}{Toys} 
    & Modified V2 & 3.1720 & 0.0248 & 1.6547\% \\ 
    & Modified V1 & 3.1530 & 0.0253 & 1.5553\% \\ 
    \bottomrule
  \end{tabular}}
\end{table}

According to the results shown in Table~\ref{tab:3}, these subtle modifications led to a significant average accuracy improvement of 1.5\% compared to the baseline model. For each dataset, the p-values from the paired t-tests were below the conventional significance threshold of 0.05, indicating that the performance differences between the extended models (Modified V1 and Modified V2) and the baseline (DiffuRec) are statistically significant across all datasets and metrics tested. Furthermore, the new variants not only outperformed the baseline but also reduced training time by approximately two-thirds, underscoring the effectiveness of our enhancements.

The comparison between the two extended models reveals that while the introduction of the cross-attention mechanism consistently improved performance across various scenarios, the impact of adding offset noise was more variable. Although offset noise occasionally enhanced performance compared to the cross-attention-only model, in some cases, the difference was minimal. Notably, extended training with offset noise did not lead to overfitting and instead promoted better generalization; however, this did not always translate into substantial performance gains.

\subsection{Overall Comparison (\textbf{RQ2})}

Table~\ref{tab:performance_comparison} presents a comprehensive comparison of our diffusion-based recommender system model, DiffuRecSys, with various conventional methods \cite{hidasi2015session, xie2022contrastive, ren2020sequential, fan2022sequential, xie2021adversarial, kang2018self, qiu2022contrastive} across three public benchmark datasets. The results demonstrate that the diffusion-based approach consistently outperforms all conventional models on both sparse datasets (e.g., Beauty, Toys) and dense dataset (e.g., ML-1M). This performance improvement is attributed to several key factors, including the use of Probabilistic Item Representations, which involves sampling from random Gaussian noise to effectively model the uncertainty of user behaviors in real-world scenarios. Additionally, by applying noise to the target item rather than the entire sequence, crucial information related to user preferences is preserved. Finally, the Diverse Inference Strategy introduces variability into the recommendation results, further enhancing overall performance.

\begin{figure}[t]
    \centering
    \caption{\small The Reconstructed Target Item Distributions after applying clustering for different users across various datasets.}
    \label{fig:GraphT}
    \includegraphics[width=1\linewidth]{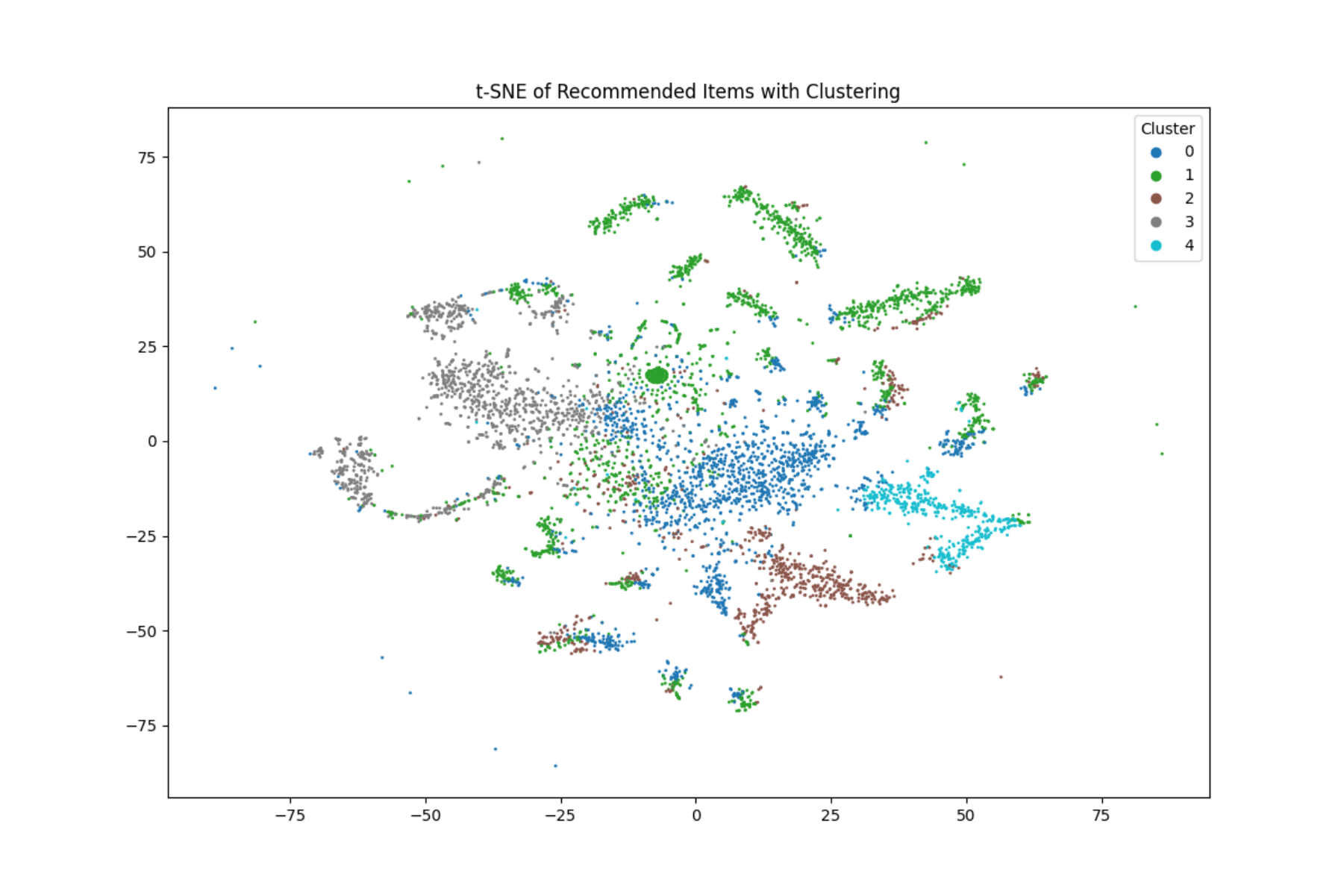}
    \caption{\small Amazon Beauty dataset}

    \centering
    \includegraphics[width=1\linewidth]{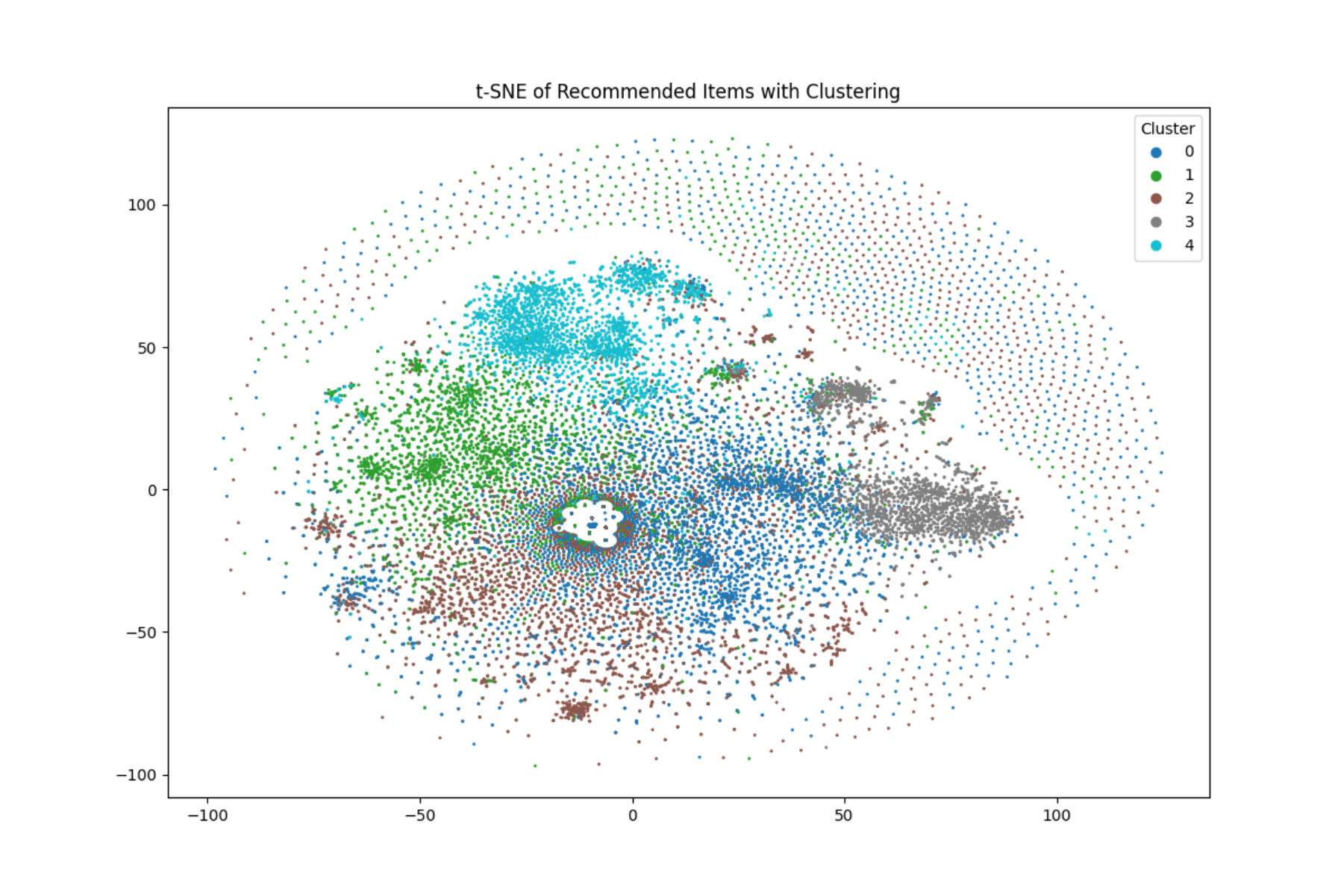}
    \caption{\small Amazon Toys dataset}

    \centering
    \includegraphics[width=1\linewidth]{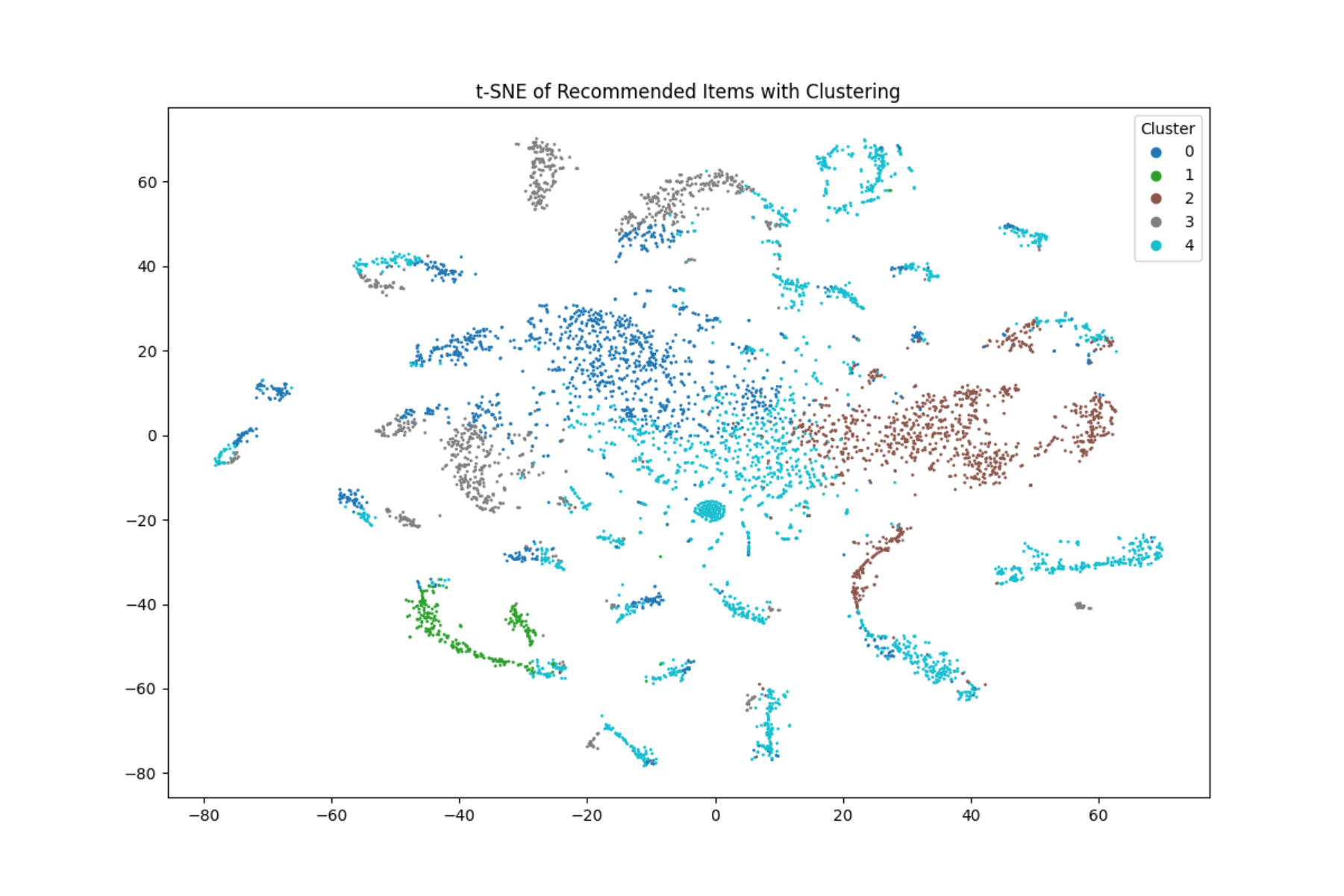}
    \caption{\small Movielens1M dataset}

\end{figure}

Figure~\ref{fig:GraphT} shows the t-distributed Stochastic Neighbor Embedding (t-SNE) visualization of the last target item for each user across different datasets, including Beauty, Toys, and ML-1M. This visualization provides insights into the distributions of the reconstructed target items across various user groups and datasets.

\begin{table*}[H]
\centering
\caption{\small Overall performance of different methods for the sequential recommendation. The best score and the second-best score in each row are bolded and underlined, respectively. The last column indicates improvements over the best baseline method.}
\resizebox{\textwidth}{!}{
\begin{tabular}{llccccccccccc}
\toprule
\textbf{Dataset} & \textbf{Metric} & \textbf{GRU4Rec} & \textbf{SASRec} & \textbf{BERT4Rec} & \textbf{STOSA} & \textbf{ACVAE} & \textbf{VSAN} & \textbf{MFGAN} & \textbf{CL4SRec} & \textbf{DuoRec} & \textbf{DiffuRecSys} & \textbf{Improv.} \\ 
\midrule
\multirow{6}{*}{Beauty} & HR@5 & 0.0206 & 0.0371 & 0.0370 & 0.0460 & 0.0438 & 0.0475 & 0.0382 & 0.0396 & \underline{0.0541} & \textbf{0.0667} & 23.29\% \\ 
 & HR@10 & 0.0332 & 0.0592 & 0.0598 & 0.0659 & 0.0690 & 0.0759 & 0.0605 & 0.0630 & \underline{0.0825} & \textbf{0.0980} & 18.79\% \\ 
 & HR@20 & 0.0526 & 0.0893 & 0.0935 & 0.0932 & 0.1059 & 0.1086 & 0.0916 & 0.0965 & \underline{0.1102} & \textbf{0.1400} & 27.04\% \\ 
 & NDCG@5 & 0.0139 & 0.0233 & 0.0233 & 0.0318 & 0.0272 & 0.0298 & 0.0254 & 0.0232 & \underline{0.0362} & \textbf{0.0458} & 26.52\% \\ 
 & NDCG@10 & 0.0175 & 0.0284 & 0.0306 & 0.0382 & 0.0354 & 0.0389 & 0.0310 & 0.0307 & \underline{0.0447} & \textbf{0.0559} & 25.06\% \\ 
 & NDCG@20 & 0.0221 & 0.0361 & 0.0391 & 0.0451 & 0.0453 & 0.0471 & 0.0405 & 0.0392 & \underline{0.0531} & \textbf{0.0665} & 25.24\% \\ \midrule
\multirow{6}{*}{Toys} & HR@5 & 0.0121 & 0.0429 & 0.0371 & \underline{0.0563} & 0.0457 & 0.0481 & 0.0395 & 0.0503 & 0.0539 & \textbf{0.0684} & 21.49\% \\ 
 & HR@10 & 0.0184 & 0.0652 & 0.0524 & \underline{0.0769} & 0.0663 & 0.0719 & 0.0641 & 0.0736 & 0.0744 & \textbf{0.0971} & 26.27\% \\ 
 & HR@20 & 0.0290 & 0.0957 & 0.0760 & 0.1006 & 0.0984 & 0.1029 & 0.0892 & 0.0990 & \underline{0.1056} & \textbf{0.1350} & 27.84\% \\ 
 & NDCG@5 & 0.0077 & 0.0248 & 0.0259 & \underline{0.0393} & 0.0291 & 0.0286 & 0.0257 & 0.0264 & 0.0340 & \textbf{0.0455} & 15.78\% \\ 
 & NDCG@10 & 0.0097 & 0.0320 & 0.0329 & \underline{0.0460} & 0.0364 & 0.0363 & 0.0328 & 0.0339 & 0.0406 & \textbf{0.0548} & 19.13\% \\ 
 & NDCG@20 & 0.0123 & 0.0397 & 0.0368 & \underline{0.0519} & 0.0432 & 0.0441 & 0.0381 & 0.0404 & 0.0472 & \textbf{0.0643} & 23.89\% \\ \midrule
\multirow{6}{*}{Movielens-1M} & HR@5 & 0.0806 & 0.1078 & 0.1308 & 0.1230 & 0.1356 & 0.1220 & 0.1275 & 0.1142 & \underline{0.1679} & \textbf{0.1957} & 16.56\% \\ 
 & HR@10 & 0.1344 & 0.1810 & 0.2219 & 0.1889 & 0.2033 & 0.2016 & 0.2086 & 0.1815 & \underline{0.2540} & \textbf{0.2843} & 11.93\% \\ 
 & HR@20 & 0.2081 & 0.2745 & 0.3354 & 0.2724 & 0.3085 & 0.3015 & 0.3166 & 0.2818 & \underline{0.3478} & \textbf{0.3958} & 13.80\% \\ 
 & NDCG@5 & 0.0475 & 0.0681 & 0.0804 & 0.0810 & 0.0837 & 0.0751 & 0.0778 & 0.0705 & \underline{0.1091} & \textbf{0.1319} & 20.90\% \\ 
 & NDCG@10 & 0.0649 & 0.0918 & 0.1097 & 0.1040 & 0.1145 & 0.1007 & 0.1040 & 0.0920 & \underline{0.1370} & \textbf{0.1605} & 17.15\% \\ 
 & NDCG@20 & 0.0834 & 0.1156 & 0.1384 & 0.1231 & 0.1392 & 0.1257 & 0.1309 & 0.1170 & \underline{0.1607} & \textbf{0.1886} & 17.36\% \\ \midrule
\end{tabular}}
\label{tab:performance_comparison}
\end{table*}

\subsection{Case Study (\textbf{RQ3})}
In line with previous studies \cite{zhang2021model, zhao2021variational, sachdeva2019sequential, fan2022sequential}, we classified the top 20\% most frequent items as head items and designated the remainder as long-tail items for performance evaluation. The model exhibits reduced performance on long-tail items compared to head items, highlighting a persistent challenge in recommending less frequent items. 

Additionally, we categorized the Amazon Beauty and MovieLens-1M datasets into five groups based on sequence length percentiles to assess model performance across different sequence lengths. In the Amazon Beauty dataset, the model generally performs better on longer sequences, as most sequences are relatively short. Conversely, in the MovieLens-1M dataset, performance gradually decreases as sequence length increases. This pattern suggests that shorter sequences may lack sufficient information to accurately predict user preferences, while excessively long sequences introduce challenges for model performance.

To further enhance recommendation performance, we propose averaging the predictions generated from different random seeds, akin to ensemble methods. This approach likely improves performance by accounting for various aspects of behavioral uncertainty through the aggregation of diverse recommendation outcomes. By leveraging this strategy, we can enhance the model's performance on very long, short, or low-frequency items. Based on these findings, we propose an efficient inference procedure, detailed in Algorithm~\ref{alg:efficient_inference}.

\begin{algorithm}
\small
\caption{\parbox[t]{0.8\columnwidth}{\small Efficient inference of $i^u_{n+1}$ for user sequence $s_u$}}
\label{alg:efficient_inference}
\KwIn{Inference sequence $\mathbf{s}_u = [\mathbf{i}^u_1, \dots, \mathbf{i}^u_n]$, sequence length $N$, diffusion steps $T$, random seeds list $\mathbf{seeds}$}
\KwOut{Average probability distribution}
\For{$i = 1$ \KwTo $|\mathbf{seeds}|$}{
    Fix random seed as $\mathbf{seeds}[i]$\;
    Sample Gaussian noise $\boldsymbol{\epsilon} \sim \mathcal{N}(0, I)$\;
    Calculate $\mathbf{x}^0_{n}$ via $q_{\phi}(\mathbf{x}_{n}^0 \mid \mathbf{i}_{n})$\;
    Calculate $\mathbf{x}^T_{n}$ via $q(\mathbf{x}^T_{n} \mid \mathbf{x}^0_{n})$\;
    Predict $\mathbf{x}^0_{N}$ via $f_{\theta}(\mathbf{x}^T_{N}, T)$\;
    Obtain distribution $\mathbf{p}_i \leftarrow \hat{p}(\mathbf{i}_{N+1} \mid \mathbf{x}^0_{N})$\;
}
\Return $\frac{1}{|\mathbf{seeds}|} \sum_{i=1}^{|\mathbf{seeds}|} \mathbf{p}_i$
\end{algorithm}

\section{Conclusion}
This work addresses the challenge of sequential recommendation systems, where existing single vector-based representations often fail to effectively capture the temporal dynamics and evolving user preferences. The introduction of diffusion models in sequential recommendation provides a more suitable framework for modeling the latent aspects of items and users' multi-level interests.

In particular, our work enhanced the Diffusion Recommender model, DiffuRec, by incorporating cross-attention mechanisms between historical user interactions and target items within the Approximator component of the model architecture. This addition allows the model to dynamically weigh historical user interactions to target items, leading to more nuanced and context-aware representations. Additionally, we introduced offset noise into the diffusion process to improve model robustness, resulting in the new model, DiffuRecSys. Another area of interest is the exploration of guided diffusion models to further improve the accuracy and relevance of user preference modeling.
 
Our experiments demonstrate the effectiveness of DiffuRecSys. However, validating its performance across a diverse range of datasets and domains is essential for establishing its generalizability. As the field of recommendation systems continues to evolve, we believe our work will inspire further innovations and applications in user-centric recommendation technologies. In conclusion, DiffuRecSys represents a novel and effective advancement in sequential recommendation, setting a new standard for intelligent and personalized recommendation systems.

%%
%% Define the bibliography file to be used

%%\bibliography{sample-2col}
%%\input{sample-2col.bbl} 

\end{document}